\documentclass[journal,twoside,web]{ieeecolor}
\usepackage{generic}
\usepackage{cite}
\usepackage{amsmath,amssymb,amsfonts}
\usepackage{algorithmic}
\usepackage{graphicx}
\usepackage{textcomp}
\usepackage{epstopdf}
\usepackage{url}

\def\BibTeX{{\rm B\kern-.05em{\sc i\kern-.025em b}\kern-.08em
    T\kern-.1667em\lower.7ex\hbox{E}\kern-.125emX}}
\begin{document}

\title{Mutually improved endoscopic image synthesis and landmark detection \\ in  unpaired image-to-image translation}

\author{Lalith Sharan, Gabriele Romano, Sven Koehler, Halvar Kelm, Matthias Karck, Raffaele De Simone, Sandy~Engelhardt

\thanks{Manuscript accepted at IEEE JBHI (Journal of Biomedical and Health Informatics). © 2021 IEEE. Personal use of this material is permitted. Permission from IEEE must be obtained for all other uses, including reprinting/republishing this material for advertising or promotional purposes, collecting new collected works for resale or redistribution to servers or lists, or reuse of any copyrighted component of this work in other works.
Manuscript received February 18, 2021, revised June 10, 2021, and July 14, 2021; accepted July 16, 2021.
This work was supported in part by Informatics for Life funded by the Klaus Tschira Foundation and the German Research Foundation DFG Project 398787259, DE 2131/2-1 and EN 1197/2-1. The authors thank Jimmy Chen for his contribution. Digital Object Identifier 10.1109/JBHI.2021.3099858}

\thanks{L. Sharan, S. Koehler, H. Kelm, and S. Engelhardt(Corresponding author), are with Dept. of Internal Med. III, Group AICM; and G. Romano, M. Karck, R. De Simone are with Dept. of Cardiac Surgery, at Heidelberg University Hospital, D-69120 Heidelberg, Germany
(email: lalithnag.sharangururaj; sven.koehler; sandy.engelhardt; gabriele.romano; matthias.karck; raffaele.de.simone@med.uni-heidelberg.de, halvar.kelm@stud.hs-mannheim.de)}}

\maketitle


\begin{abstract}

The CycleGAN framework allows for unsupervised image-to-image translation of unpaired data. In a scenario of surgical training on a physical surgical simulator, this method can be used to transform endoscopic images of phantoms into images which more closely resemble the intra-operative appearance of the same surgical target structure. This can be viewed as a novel augmented reality approach, which we coined  \textit{Hyperrealism} in previous work. In this use case, it is of paramount importance to display objects like needles, sutures or instruments consistent in both domains while altering the style to a more tissue-like appearance. Segmentation of these objects would allow for a direct transfer, however, contouring of these, partly tiny and thin foreground objects is cumbersome and perhaps inaccurate. 
Instead, we propose to use landmark detection on the points when sutures pass into the tissue. This objective is directly incorporated into a CycleGAN framework by treating the performance of pre-trained detector models as an additional optimization goal. 
We show that a task defined on these sparse landmark labels improves consistency of synthesis by the generator network in both domains.
Comparing a baseline CycleGAN architecture to our proposed extension (\textit{DetCycleGAN}), mean precision (PPV) improved by $+61.32$, mean sensitivity (TPR) by $+37.91$, and mean $F_1$ score by $+0.4743$.
Furthermore, it could be shown that by dataset fusion, generated intra-operative images can be leveraged as additional training data for the detection network itself. 
The data is released within the scope of the AdaptOR MICCAI Challenge 2021 at \url{https://adaptor2021.github.io/}, and code at \url{https://github.com/Cardio-AI/detcyclegan_pytorch}.

\end{abstract}



\section{Introduction}
\label{sec:Introduction}

\begin{figure}[t]
    \centering
    \includegraphics[width=0.5\textwidth]{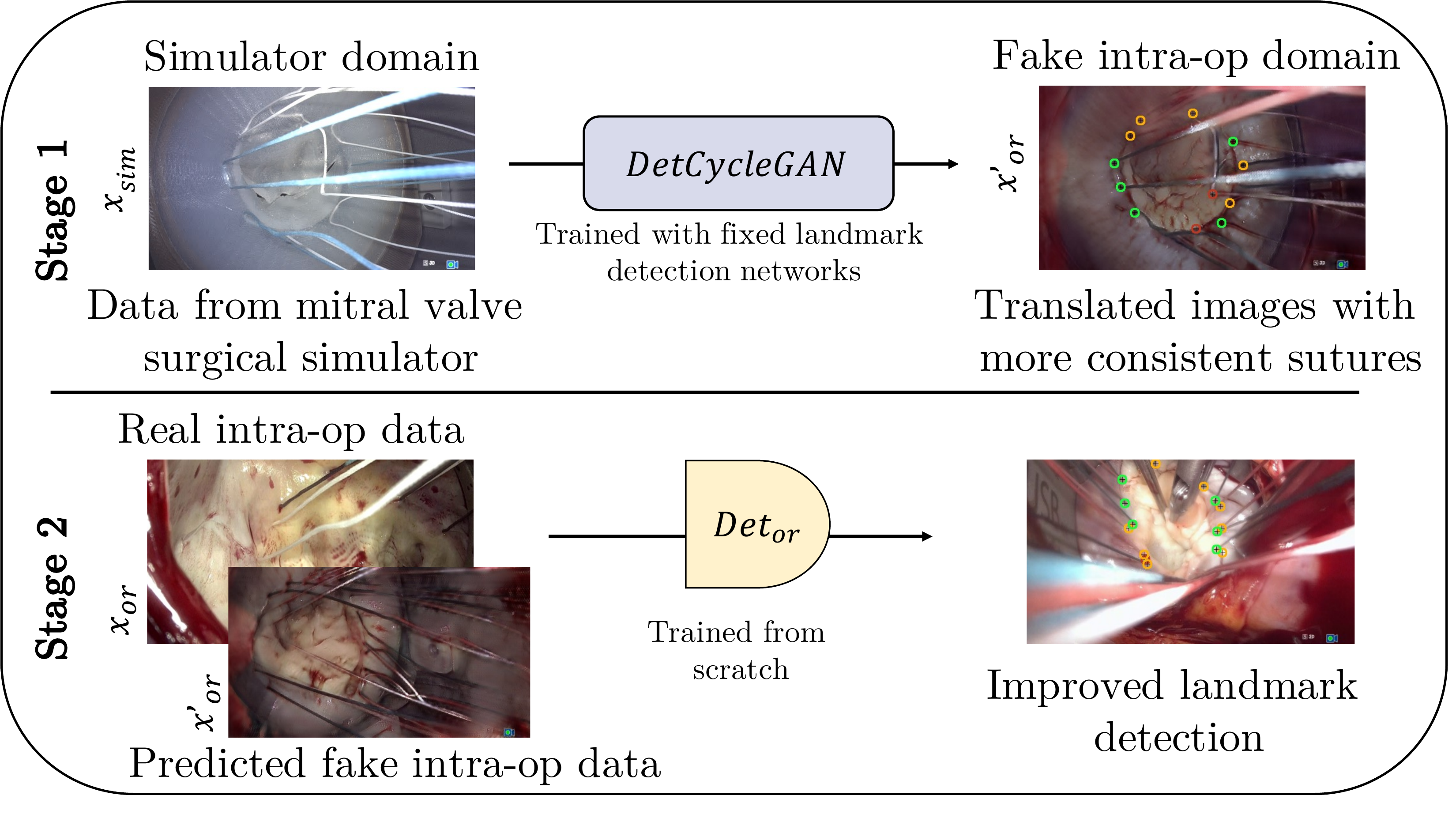}
     \caption{An overview of the workflow, for mutually improved endoscopic image synthesis (Stage $1$), and landmark detection (Stage $2$).}
    \label{fig:workflow}
\end{figure}

Mitral valve repair is a technique to treat mitral regurgitation, that is increasingly performed in a minimally invasive setup \cite{casselman_filip_p._mitral_2003}. It is a complex surgery with a steep learning curve, demanding a high level of skill and dexterity from the surgeon. Surgical simulators with patient-specific valve replicas are one of the solutions to provide a tool for surgical education, training, and planning for mitral valve repair \cite{Engelhardt_IJCARS19}. However, depending on the material used, the optical appearance of such phantom simulators may be in stark contrast to that of the intra-operative tissue. 

Enhancing the realism of surgical simulators is a pertinent problem to improve the perception of the trainee, and has an impact on the quality of surgical training. Our previous works \cite{Engelhardt_Miccai18, engelhardt_cross-domain_2019}, lay a foundation based on \textit{Generative Adversarial Networks (GAN)} \cite{goodfellow_generative_2014}, to transform the images acquired during surgical training (source domain), to realistic images that closely resemble the intra-operatively acquired images (target domain). This approach, coined \textit{Hyperrealism} is a new augmented reality paradigm, for minimally-invasive surgical training with physical models \cite{Engelhardt_Miccai18}. \textit{Hyperrealism}, in combination with quantitative analysis of endoscopic data forms the basis of a toolkit for preparation, planning and guidance of complex surgeries such as mitral valve repair. One such quantification task is the detection of the positions when sutures are entering or exiting the tissue from the endoscopic images \cite{DBLP:conf/bildmed/SternSRKKSWE21}, for example, to compute the number of used sutures or the distances between them. In mitral valve repair, suturing is performed among others during annuloplasty, which involves anchoring of a prosthetic ring to stabilize or downsize the pathological valve.  

Generative adversarial networks based on the CycleGAN \cite{zhu_unpaired_2020} model have shown great promise in the recent years in the field of computer vision \cite{shamsolmoali_image_2021}, in particular for unpaired image-to-image translation applications. However, they are prone to flickering on videos, are challenging to train, and perform poorly when there is high heterogeneity between the source and target domains \cite{DBLP:journals/corr/abs-2101-08629}. Moreover, image translation across highly dissimilar domains can lead to unwanted feature hallucination \cite{cohen_distribution_2018}. Our previous works \cite{Engelhardt_Miccai18, engelhardt_cross-domain_2019}, address some of these issues with temporal and cross-domain stereoscopic constraints. Nevertheless, it is still a challenge to produce physically consistent images in the target domain. For example, the appearance of objects such as sutures, needles, instruments, and prostheses that are already realistic in the simulator domain, should be consistent when transformed to the intra-operative domain.

Several approaches have been proposed to improve the quality of image-to-image translation, e.g. by integrating information from auxiliary tasks such as object detection \cite{yu_unsupervised_2020, bhattacharjee_dunit_2020, arruda_cross-domain_2019}, facial landmark detection \cite{wu_landmark_2019, piao_semi-supervised_2019}, or semantic segmentation \cite{hoffman_cycada_2017, DBLP:conf/iclr/LeeRLG19}. However, some of these methods are ill-suited to address the problem in the endoscopic domain, in particular for mitral valve repair. 
While object annotations from traffic scenes or labels of facial landmarks are widely available from public datasets \cite{ros_synthia_2016, menze_object_2015}, surgical datasets are difficult to acquire, underlie data privacy restrictions and are cumbersome to annotate. The challenging nature of the datasets complicate the matter, as they are typically marred with endoscopic artefacts such as fogging, occlusions, tissue specularities and reflection from surgical instruments.
Synthetic and real domains in the field of autonomous driving, indoor object detection, and facial data, contain relatively more structural consistency with respect to the landmarks, in comparison with the simulator and intra-operative domains in mitral valve repair.
The suture detection task in mitral valve repair involves identifying an unknown number of suture points, where the sutures lack a clear structural relation, suffer from self-occlusion or are occluded by other objects, thereby rendering suture detection a challenging task in itself.

In this work, we propose a multi-stage procedure (cf. Figure \ref{fig:workflow}) to mutually improve the performance of unpaired image-to-image translation \textit{and} suture detection with a \textit{Detection-integrated CycleGAN (DetCycleGAN)} architecture in a very challenging domain. 
We use a fixed suture detection network \cite{Sharan_IJCARS} integrated with a CycleGAN network, to first produce more physically plausible image translation. This translation generates frames from the intra-operative domain conditioned on the simulator domain and vice versa. 
In the next stage, a new suture detection network is trained from scratch with a combined dataset comprising the generated, more-consistent, fake intra-operative images and the real intra-operative data. Using a fused dataset with real and fake data improves the performance of the suture detection network without the need for additional intra-operative data. 
Besides the improved realism of generated images in stage $1$, the output of stage $2$ can potentially facilitate quantification, improved documentation, and implementation of augmented reality overlays \cite{Engelhardt_MICCAI2014, Engelhardt_SPIE2016}.

\section{Related Work}
\label{sub:related_work}

Generative adversarial networks (\textit{GANs}) introduced by Goodfellow \textit{et al.} \cite{goodfellow_generative_2014}, have made vital contributions in the field of computer vision \cite{toldo_unsupervised_2020, DBLP:journals/corr/abs-2101-08629, yi_generative_2019, shamsolmoali_image_2021}. Generative models have been employed to tackle a wide range of applications spanning image synthesis, image translation and style transfer \cite{isola_image--image_2018}, \cite{zhu_unpaired_2020}, \cite{yi_dualgan_2018}, $3D$ reconstruction and shape completion \cite{wang_shape_2017}, \cite{hermoza_3d_2018}, and adversarial domain adaptation \cite{hoffman_cycada_2017}, \cite{chen_crdoco_2020}, \cite{DBLP:conf/iclr/LeeRLG19}. 

Fundamentally, \textit{GANs} comprise of two competing networks, a Generator $G$ and a Discriminator $D$ that work in conjunction to estimate the source data distribution. More formally, assuming a set of data samples $x$ arise from a source distribution $p_s$, a generator network $G$
models a mapping $G: z\xrightarrow{}x_g$ from Gaussian noise samples $z$ to generated samples $x_g$. A discriminator network $D$ takes as input the generated images $G(z)$ and $x$, and distinguishes between the real and fake images. Together, the networks play a min-max game to learn the distribution ${\hat{p}_s}$ that approximates $p_s$.

\subsection{Image-to-image translation}
\label{sub:image_to_image_translation}
\textit{Conditional GANs} or \textit{cGANs}, are models that are conditioned on the input data, instead of a noise distribution. \textit{cGAN} models \cite{isola_image--image_2018} are well-suited for image-to-image translation tasks, where the objective is to learn a mapping from the source to target domain while preserving the semantic content of the source domain. Multi-modal translation networks \cite{zhu_toward_2017}, \cite{DBLP:conf/nips/DentonB17} learn to transform the input from the source domain to multiple target domains or styles. Unpaired image-to-image translation methods \cite{yi_dualgan_2018}, \cite{zhu_unpaired_2020}, \cite{liu_unsupervised_2017}, \cite{li_unsupervised_2018} map the source and target domains without access to paired data. The authors of \cite{yi_dualgan_2018} and \cite{zhu_unpaired_2020} use cycle consistency loss by transforming the mapped image back to the source domain, in addition to the adversarial losses. \textit{UNIT} proposed by Liu \textit{et al.} \cite{liu_unsupervised_2017} models the mapping with a VAE-GAN, making a shared latent space assumption. \textit{SCANs} from Li \textit{et al.} \cite{li_unsupervised_2018} proposes multi-stage cycle-consistent learning. Enforcing cycle-consistency helps to learn a source-to-target transformation despite the lack of paired data. 
However, despite their great performances, it is known that the translation networks often only learn the coarse joint distribution and style content of the two domains, compromising the consistency of fine-grained information \cite{DBLP:journals/corr/abs-2101-08629}.

\subsection{Landmark Detection}
\label{sub:landmark_detection_rel_work}
Landmark detection is a task which deals with localization of a sparse set of points in an image, that commonly finds applications in pose estimation, image registration and augmented reality-based visualisations. Existing discriminative deep learning-based approaches to landmark detection broadly fall under three categories. 
Regression based approaches estimate a set of coordinates directly from the image space \cite{yan_survey_2018,fan_approaching_2016,yang_stacked_2017}. These methods require a fixed number of landmark locations which are then regressed. In contrast, heatmap-based approaches estimate the likelihood distribution around the landmark coordinates \cite{payer_integrating_2019, Chandran2020AttentionDrivenCF}. Graph-based landmark detection methods \cite{zhou_exemplar-based_2013}, \cite{vedaldi_structured_2020}, use a graph to model both the positions of the landmark points and the relationship between them.

In our case, the landmark detection task involves identifying an unknown number of suture entry and exit points. In our earlier work \cite{Engelhardt_MICCAI2014}, we used a random forest approach to detect these suture points and exploited them as markers for augmented reality visualizations. In a follow-up work \cite{Engelhardt_SPIE2016}, these points were temporally tracked using optical flow. Our previous work  \cite{DBLP:conf/bildmed/SternSRKKSWE21} demonstrates a  heatmap-based localisation model based on the U-Net \cite{Ronneberger_Unet2015} architecture to tackle this problem. Two models, one each for the simulator and the intra-operative domain were trained on two respective datasets from these domains. In our recent work \cite{Sharan_IJCARS}, we extend this architecture with a differentiable \textit{Gaussian} filter and spatial \textit{Soft-Argmax} layer to produce more stable results in the simulator and intra-operative domains. In this work, this architecture is used for the combined experiments with the image-translation methods.

\subsection{Task-integrated image-to-image translation}
\label{sub:task_integrated_image_to_image_translation}
Multiple approaches from the literature propose combining CycleGANs with task networks to tackle various applications such as facial transformations, and synthetic to real transformation of driving scenes.
Wang \textit{et al.} \cite{wang_thermal_2018} incorporates a detector network to produce identity-preserving facial transformations. Here, a single detector is used across both the domains and is trained together with adversarial networks. Wu \textit{et al.} \cite{wu_landmark_2019} use facial landmarks, with a landmark consistency loss, to produce structure-preserving facial transformations. The facial landmarks, are concatenated with the image from the source domain, and provided to the generator. Additionally, global and local discriminators based on the landmarks are used. Ha \textit{et al.} \cite{ha_marionette_2019} used a landmark transformation block for $3D$ facial landmarks for identity-preserving facial reenactment. 
Piao \textit{et al.} \cite{piao_semi-supervised_2019} jointly optimise a CycleGAN with a shape reconstruction network. In these approaches, however, a single landmark detector network is used for both the source and target domains, to perform a combined training with the GAN. Owing to the considerable domain gap in our use-case, our method uses two individual pre-trained detectors that can deal with a varying number of landmarks per image.
In the area of autonomous driving and traffic scenes, task networks for object detection are often combined with a CycleGAN to impart semantic cues and improve image translation. Lin \textit{et al.} \cite{lin_multimodal_2020} train a multi-modal CycleGAN together with segmentation maps from traffic scenes, to generate structure-consistent translations across domains. In AugGan \cite{ferrari_auggan_2018}, the authors use a CycleGAN trained together with a segmentation sub-task to produce structure-aware image translation for object detection on traffic datasets. \textit{DUNIT} \cite{bhattacharjee_dunit_2020} provides object instances together with a global image to the generator for better image translation. Shao \textit{et al.} \cite{shao_feature_2021} use a feature-enhancement module trained together with a CycleGAN to improve performance on vehicle object detection. These approaches involve using a bounding box network or a dense semantic segmentation map as the task network.

Another objective of coupling image-to-image translation with task networks is to translate data from the source domain to the target domain towards the goal of closing the domain gap and improving performance of a task network in the target domain. This task can be approached through \textit{semi-supervised learning} \cite{van_engelen_survey_2020, zhou_collaborative_2019} which typically uses a set of labeled data, together with a dataset with pseudo labels or no labels. Zhang \textit{et. al} \cite{zhang_semi-supervised_2020}, Kumar \textit{et al.} \cite{kumar_s2ld_2020}, and Honari \textit{et al.} \cite{DBLP:conf/cvpr/HonariMTVPK18} use semi-supervised learning for landmark localization. This forms an important part of a larger class of problems belonging to the task of domain adaptation \cite{toldo_unsupervised_2020}. Here, the objective can be formulated as follows: given a set of samples $x_S$ belonging to the source domain $X_s$ with labels $y_s$, the objective is to estimate the labels $y_t$, for samples $x_t$ belonging to the target domain distribution $X_t$. Multiple methods have been proposed over the recent years to tackle the problem of domain adaptation \cite{li_bidirectional_2019}, \cite{chen_crdoco_2020}, \cite{hoffman_cycada_2017}, \cite{wu_dcan_2018}, \cite{zhang_curriculum_2017}. CyCADA \cite{hoffman_cycada_2017} proposes a domain translation approach with a semantic segmentation task with added semantic constraints for image translation. In Chen \textit{et al.} \cite{chen_crdoco_2020}, a cross-domain perceptual loss is applied between the source and the translated source domains. 
Li et al. \cite{li_bidirectional_2019} propose a bidirectional learning framework to iteratively improve the tasks of image translation and semantic segmentation, in the absence of labels in the target domain.

However, in contrast to the task of unsupervised domain adaptation using semi-supervised learning, like in CyCADA \cite{hoffman_cycada_2017}, the objective of our work is not to learn a task network in the target domain. In this work, we make use of the intra-operative dataset and the landmark detection labels from both the simulator and the intra-operative domains, to further mutually improve the performance of the image translation task and the landmark detection task. Furthermore, in our case the landmark detection task which involves detecting sparse and varying amount of points, is different and more challenging in comparison with classification or semantic segmentation task networks typically used in the previously mentioned applications.

In contrast to existing work, this work proposes the following contributions. 
\begin{itemize}
\item The proposed multi-stage framework (Figure \ref{fig:workflow}) has a two-fold objective:
    \begin{itemize}
    \item Firstly, fixed landmark detection networks are integrated with a CycleGAN, termed \textit{DetCycleGAN}, to produce more realistic and  physically plausible suture-consistent image translations from the simulator to the intra-operative domain. 
    \item Secondly, we in turn train the landmark detection network by fusing real and fake intra-operative images and show a performance improvement over the original detector network.
    This means that data from the simulator domain, when translated to the intra-operative domain appropriately, is useful to improve the performance of landmark detection.  
    \end{itemize}
\item  Manual foreground contouring is time intensive. We show in a tailored analysis that less labor intensive annotation tasks like landmark labelling are suited to be used as ground truth to improve the consistency of the whole foreground structure, in this case, sutures. 
\end{itemize}

\section{Material and Methods}
\label{sec:material_and_methods}

\begin{figure}[t]
    \centering
    \includegraphics[width=0.5\textwidth]{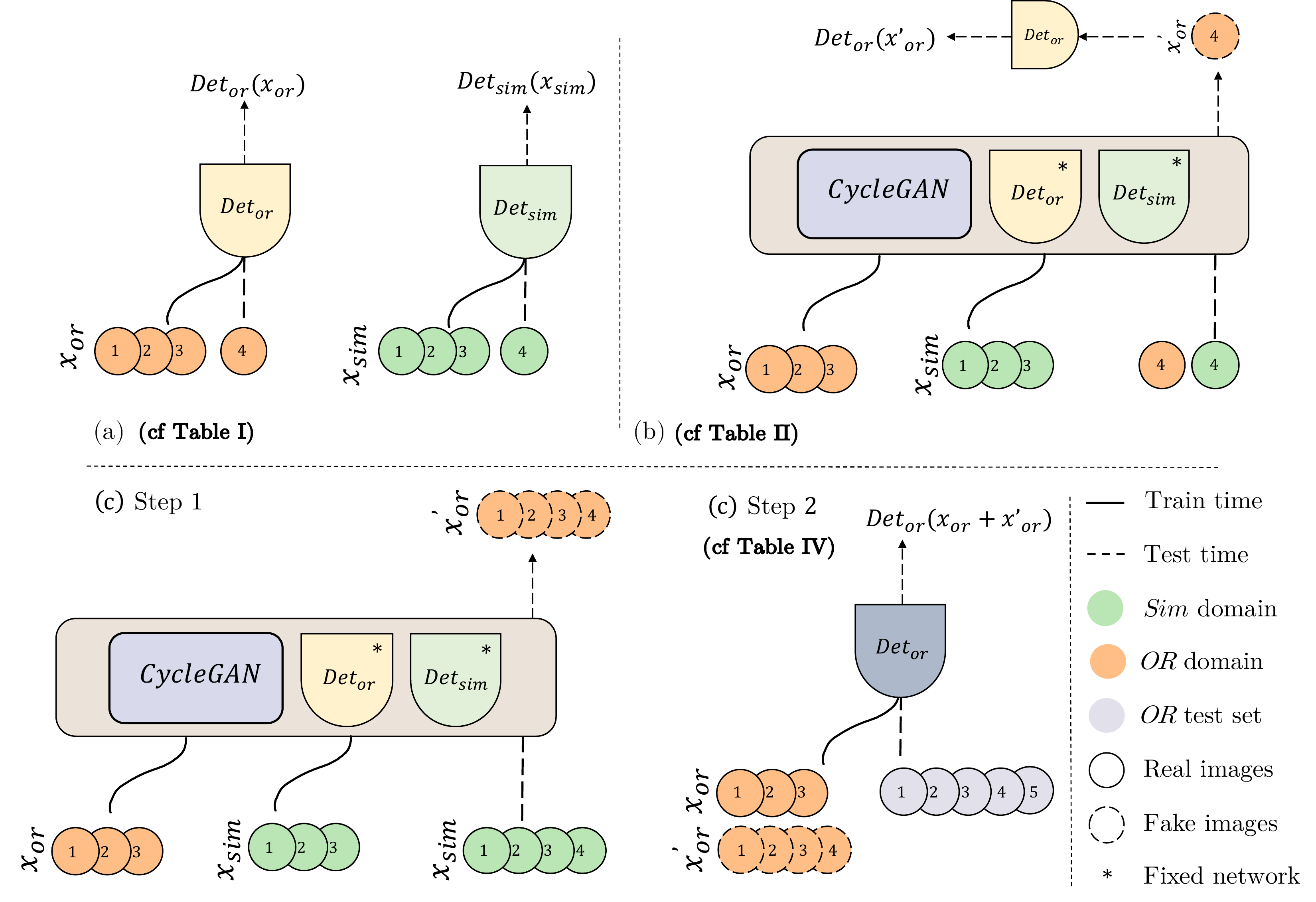}
    \caption{An overview of the multi-stage method together with the cross-validation evaluation strategy. The numbers in the circles indicate the different cross-validation folds. (a) The landmark detection networks $Det_{or}$ and $Det_{sim}$ are first trained on the respective domains to detect suture landmarks. (b) An unpaired image-to-image translation network makes use of the learnt information (fixed task networks) to produce suture-consistent image translation from simulator domain to intra-operative and vice versa. (c) From the 4 $G_{sim2or}$ generators trained in the previous step, fake $x'_{or}$ data is synthesized and fused with the intra-operative data to train $Det_{or}$. The final performance is assessed on a completely independent test set.}
    \label{fig:process}
\end{figure}

The study was approved by the Local Ethics Commitee from University Hospital Heidelberg. Registration numbers are S-658/2016 (12.09.2017) and S-777/2019 (20.11.2019).
An overview of the proposed method to transform endoscopic images from the simulator to the intra-operative domain with consistent suture landmarks, is illustrated in Figure \ref{fig:process}. Firstly, a heatmap-based landmark detection network learns to detect the suture landmarks from the endoscopic images. Two separate networks with the same underlying architecture are trained, for the simulator and the intra-operative domain respectively (Figure \ref{fig:process}a). In the first stage, an unpaired image-to-image translation network is trained in a framework that directly incorporates the pre-trained landmark detection networks to improve image synthesis. The image translation networks learn to translate the endoscopic images from the simulator domain to the intra-operative domain and vice versa. In the prediction step, a set of images from the simulator domain is transformed to the intra-operative domain through the image translation network and the landmark detection network from Figure \ref{fig:process}a is used to detect the landmarks from the transformed images (Figure \ref{fig:process}b). For the second stage, the generated fake data is utilized to extend the current training set for the intra-operative detector net to in turn improve performance of the task network itself (Figure \ref{fig:process}c). The problem formulation, and a breakdown of each network component is presented in the remainder of this section.

\subsection{Landmark detection network} 
\label{sub:landmark_detection_methods}
The goal of the landmark detection task is to learn the locations of the suture landmarks from endoscopic images $x_{dom}$ in the simulator and the intra-operative domain, $dom \in \{sim, or\}$, given a set of point labels. Let $x_{dom} \in \mathbb{R}^{W\times H \times3}$ be a $3$-channel RGB image. $H$ and $W$ are the height and width of the image respectively. Let $y_{dom}$ denote the ground truth label image corresponding to the image $x_{dom}$, such that $y_i \in \{0, 1\}$ for $i= 1,..., W\times H$. The objective is to estimate labels $\hat{y}_{dom}$ of an image $x_{dom}$, with a landmark detection network $Det_{dom}$.

The number and relative orientation of the suture landmarks varies from image to image, besides variations in the scale, illumination and view of the scene. Moreover, the entry and exit points of the sutures, often appear close to one another. Therefore, landmark detection for this use case is particularly challenging. 
We use an extension of our previous architecture \cite{DBLP:conf/bildmed/SternSRKKSWE21, Sharan_IJCARS} for suture point detection and formulate the problem as a multi-instance sparse heatmap-based landmark regression task. The network $Det_{dom}$ is a U-Net \cite{Ronneberger_Unet2015} based encoder-decoder architecture, with a differentiable \textit{Gaussian} filter and a convolutional $2D$ spatial \textit{Soft-Argmax} layer.
A $2D$ Gaussian heatmap is computed centered on each ground truth landmark in $y_{dom}$. For each image $x_{dom}$, the heatmap image with pixel labels ${y_i} \in [0, 1]$ is fed as input to the landmark detection model. 
The network is trained by optimizing \textit{point segmentation losses} at two different stages, each of which is a combination of the \textit{Mean Squared Error (MSE)} loss $\mathcal{L}_{MSE}$ and the \textit{Soft Dice loss} $\mathcal{L}_{Dice}$,

\begin{equation}
\mathcal{L}_{Det} = \sum_{i=1}^{H \times W}  (y_i - \hat{y}_i)^2 + 1 - 
{\frac {2{\sum_{i=1}^{H \times W} y_i  \hat{y}_i} + {s}} 
{{\sum_{i=1}^{H \times W} y_i} + {\sum_{i=1}^{H \times W} \hat{y}_i}+s}}
\label{eq:seg_loss}
\end{equation}

 where $s$ is a commonly used smoothing factor. 

\subsection{Image translation network} 
\label{sub:image_translation}
The unpaired image-to-image translation network is a generative adversarial network \cite{goodfellow_generative_2014} based on the CycleGAN \cite{zhu_unpaired_2020} architecture. Figure \ref{fig:architecture} illustrates the architecture of the network. The network takes as input, sets of unpaired images from the simulator and intra-operative domains, and through adversarial learning generates images closer to the target domain. For each of the domains, we can define a generator $G$, that performs a mapping between two domains. Let $G_{sim2or}$ be the generator function $G_{sim2or}: x_{sim}\rightarrow x'_{or}$, where $x_{sim} \in X_{sim}$ make up the set of images in the simulator domain, and the images transformed to the intra-operative domain are denoted by $x'_{or}$. Similarly the generator $G_{or2sim}: x_{or}\rightarrow x'_{sim}$ transforms images from the intra-operative to the simulator domain. The adversarial discriminators $D_{dom}$, one for each domain $dom \in \{sim, or\}$, take as input the images from the true or from the translated domain and classifies them as real and fake, i.e.\ $D_{dom}: x_{dom}\rightarrow 1$ and $D_{dom}: x_{dom}'\rightarrow 0$. The goal of the GAN training is to fool the discriminator \cite{zhu_unpaired_2020}. Therefore, adversarial losses to both mapping functions are applied. For the mapping function $G_{sim2or}$ and its discriminator $D_{or}$, we express the objective as

\begin{equation}
    \begin{aligned}
        \mathcal{L}_{GAN} &= \mathbb{E}_{x_{or} \sim X_{or}} [\log(D_{or}(x_{or})] \\
        &+ \mathbb{E}_{x_{sim} \sim X_{sim}} [\log(1 - D_{or}(x'_{or})].
    \end{aligned}
\label{eq:gan_loss}
\end{equation}
The respective adversarial loss  for the mapping function $G_{or2sim}$ and its discriminator $D_{sim}$ is also applied.

The translated image $x'_{or}$, is further passed to the corresponding generator $G_{or2sim}$ to get the recovered output in the original domain. A cycle consistency loss $\mathcal{L}_{cycle}$ is then imposed on the recovered images, thereby enforcing the network to reproduce the images faithfully,

\begin{equation}
    \begin{aligned}
        \mathcal{L}_{cycle} &= \mathbb{E}_{x_{or} \sim X_{or}} [\| x''_{or} - x_{or} \|] \\
                &+ \mathbb{E}_{x_{sim} \sim X_{sim}} [\| x''_{sim} - x_{sim} \|].
    \end{aligned}
\label{eq:cycle_loss}
\end{equation}

Additionally, an identity loss  $\mathcal{L}_{ident}$ is imposed as regularization of the generator network to be near an identity mapping when samples from the target domain are provided as input \cite{DBLP:conf/iclr/TaigmanPW17}. This loss is defined as

\begin{equation}
    \begin{aligned}
        \mathcal{L}_{ident} &= \mathbb{E}_{x_{or} \sim X_{or}} [\| {G_{sim2or}({x_{or}}) - {x_{or}}} \|] 
        \\
        &+ 
                \mathbb{E}_{x_{sim} \sim X_{sim}} [\| {G_{or2sim}({x_{sim}}) - {x_{sim}}} \|].
    \end{aligned}
\label{eq:ident_loss}
\end{equation}

This preserves color composition in a more reliable way, as shown by Zhu \textit{et al.} \cite{zhu_unpaired_2020}. The unpaired image-to-image translation networks are trained by optimizing the objective given by

\newcommand{\Gsimor}{G_{sim2or}}
\newcommand{\Gorsim}{G_{or2sim}}

\begin{equation}
    \begin{aligned}
        \mathcal{L}_{CycleGAN} &= \min_{\substack{\Gsimor\\ \Gorsim}} \max_{\substack{D_{sim}\\ D_{or}}} [ \mathcal{L}_{GAN}(G_{sim2or}, D_{or})\\
        &+ 
         \mathcal{L}_{GAN}(G_{or2sim}, D_{sim})
        + 
        \lambda_1 \mathcal{L}_{cycle} 
        + 
        \lambda_2 \mathcal{L}_{ident}]
    \end{aligned}
\label{eq:CycleGAN}
\end{equation}

where  
$\lambda_1 = 10$ and $\lambda_2 = 5$ are weights to balance the loss, which are set empirically.

\subsection{Combined task formulation} 
\label{sub:combined_task}
The image translation network presented in \ref{sub:image_translation} learns the underlying distribution of the data in order to transform the images from the source to target domain and vice versa. However, the translation is performed in a manner that does not necessarily emphasize semantic consistency. The network learns features that are sufficiently discriminative of the two domains, although they may be semantically irrelevant. Moreover, the network has problems learning fine-grained information. In order to produce a more semantically consistent transformation, we directly incorporate a landmark detection network $Det_{dom}$ into the CycleGAN framework. Detection is performed on the transformed $x_{dom'}$ and back-transformed $x_{dom''}$ images. 
In the original CycleGAN method, the result of image synthesis is evaluated by the discriminator networks. However, we extend on this by evaluating whether the position of the suture points are at the same location as in the source domain. The error on the detection rate is minimized during GAN optimization and backpropagated into the generator network. 

\begin{figure*}
    \centering
    \includegraphics[width=0.9\textwidth]{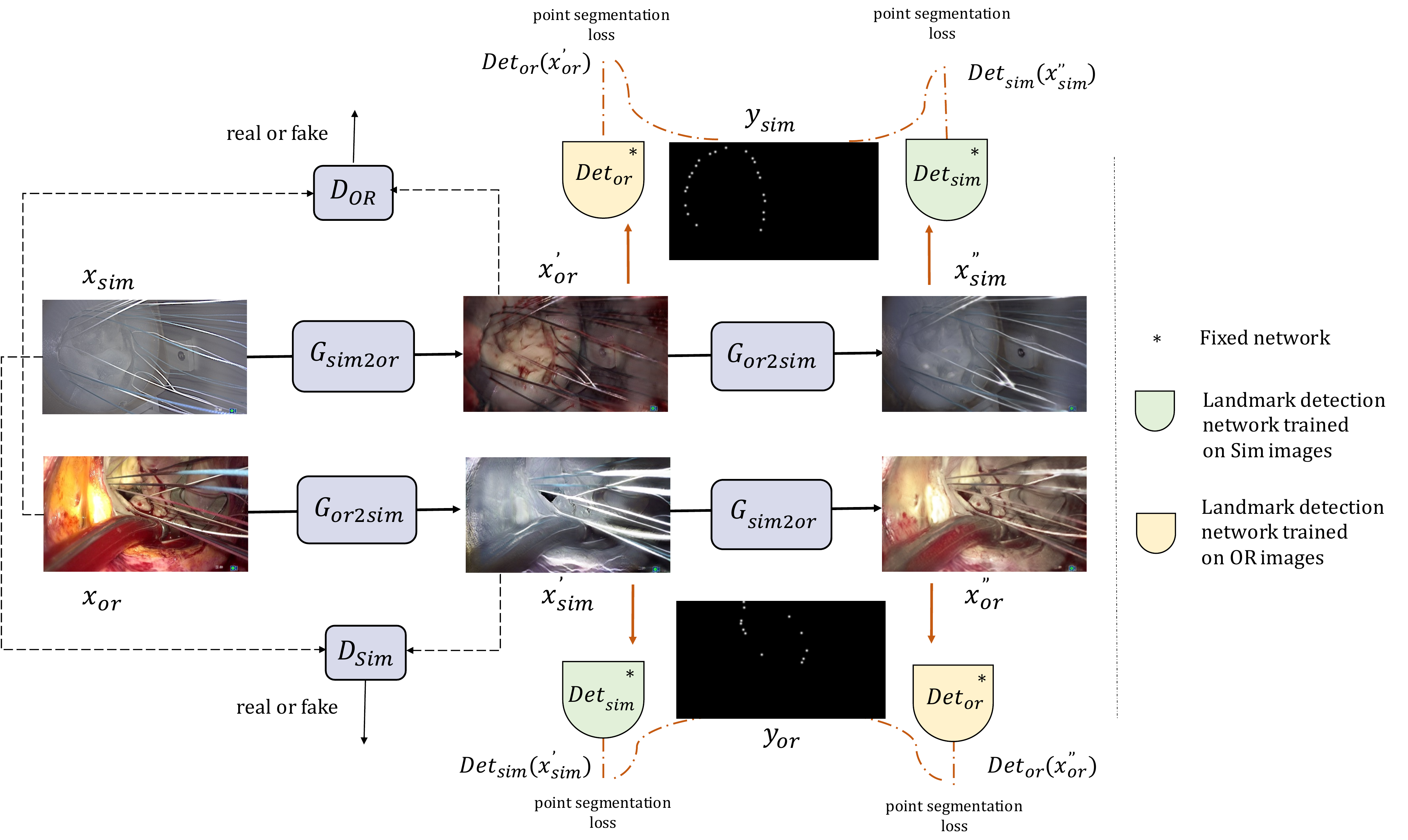}
    \caption{The proposed \textit{DetCycleGAN} architecture. The unpaired image-to-image translation network operating on the simulator and the intra-operative domains (with black lines), and the proposed integration of fixed landmark detection networks (with red lines) are shown.}
    \label{fig:architecture}
\end{figure*}

In the following, we will describe the GAN extension for the mapping function $G_{sim2or}: x_{sim} \rightarrow x_{or'}$, however, the mapping in the other direction follows accordingly. The novel part is indicated by the red lines in Figure \ref{fig:architecture}.
For every image $x_{sim}$, the corresponding transformed image $x'_{or}$ is fed to the pre-trained suture landmark detection network $Det_{or}$ with fixed weights, to obtain the corresponding predicted labels. The predicted labels $Det_{or}(x'_{or})$ are enforced to be consistent with the ground truth labels $y_{sim}$ through the \textit{detection consistency loss},

\begin{equation}
    \begin{aligned}
        \mathcal{L}_{Det'} &= \mathcal{L}_{Det}(Det_{or}(x'_{or}), y_{sim}) \\ 
        &+  \mathcal{L}_{Det}(Det_{sim}(x'_{sim}), y_{or}).
    \end{aligned}
\label{eq:point_loss}
\end{equation}

Similarly the recovered images $x''_{sim}$ from the simulator domain are passed through the pre-trained landmark detection network $Det_{sim}$ to obtain predicted labels $Det_{sim}(x''_{sim})$. The labels are enforced to be consistent with the corresponding ground-truth labels ${y}_{sim}$ by the loss, 

\begin{equation}
    \begin{aligned}
        \mathcal{L}_{Det''} &= \mathcal{L}_{Det}(Det_{sim}(x''_{sim}), y_{sim}) \\ &+  \mathcal{L}_{Det}(Det_{or}(x''_{or}), y_{or}).
    \end{aligned}
\label{eq:point_rec_loss}
\end{equation}

The unpaired image-to-image translation network, is trained by making use of the learnt landmarks in both the domains, by optimizing the final joint objective, given by 

\begin{equation}
    \begin{aligned}
        \mathcal{L}_{DetCycleGAN} &= \mathcal{L}_{CycleGAN} +  \alpha_1 \mathcal{L}_{Det'}+  \alpha_2 \mathcal{L}_{Det''}.
    \end{aligned}
\label{eq:final_objective}
\end{equation}

where $\alpha_1$ and $\alpha_2$ are weighting factors. In this work, we propose and evaluate two model variants. In Variant $1$, both $\mathcal{L}_{Det'}$ and $\mathcal{L}_{Det''}$ are used ($\alpha_1, \alpha_2 >0$), and in Variant $2$, only $\mathcal{L}_{Det'}$ is used ($\alpha_2=0$).

\section{Experiments}
\label{sec:experiments}

\subsection{Datasets} 
\label{sub:datasets}
The simulator and intra-operative datasets used in this work were acquired in the department of \textit{Cardiac Surgery}, at the \textit{University Hospital Heidelberg, Germany}. The \textit{Image 1S 3D}, $30^{\circ}$ degree optics stereo-endoscope (Karl Storz SE \& Co. KG, Tuttlingen, Germany) was used for imaging. 

\subsubsection{Simulator (\textit{sim}) domain}
\label{subsub:simulator_domain}
A surgical simulator, introduced in Engelhardt \textit{et al.} \cite{Engelhardt_IJCARS19}, is used towards the goal of surgical planning and training for mitral valve repair surgeries. The simulator houses patient-specific mitral valve replicas, from which endoscopic video is captured during a surgical simulation, at different illuminations, depth, pose, number of objects in the scene, etc. The \textit{Image1 Connect TC200} (Karl Storz SE \& Co. KG, Tuttlingen, Germany) was used to acquire images from the stereo-endoscope at a resolution of $1920 \times 1080$, in top-down format with the two halves corresponding to the left and the right images of the stereo-pair respectively. Two recording systems were used for the data capture, namely: the \textit{AIDA} (Karl Storz SE \& Co. KG, Tuttlingen, Germany) to capture at $25 fps$, and the \textit{DVI2PCIe} video capture card (Epiphan Video, California, USA) to capture at $1 fps$. The relevant frames were then extracted from the captured video, with the sampling performed every $240th$, $120th$ or $10th$ frame depending on the amount of motion in the scene. The dataset is formed from the extracted frames, comprising $2,708$ images ($1,354$ stereo-pairs) from $10$ simulator sessions. For both the simulator and the intra-operative domains, the left and the right images of the stereo-pair are treated separately for this work. For the images in the dataset, the entry and exit points of the sutures were annotated manually using the \textit{labelme} annotation tool. A total of approx. $33,800$ suture landmarks were labelled from images in the simulator domain. The prepared simulator dataset is organized into $4$ folds.

\subsubsection{Intra-operative (\textit{OR}) domain}
\label{subsub:or_domain}
In the intra-operative domain, images were acquired from minimally invasive mitral valve repair surgeries.  The heterogeneous dataset comprises of scenes with varying illuminations, white balance, pose, viewing angle and endoscopic artefacts. Two imaging systems were used for the capture of the endoscopic data, namely: the \textit{Image1 Connect TC200} (Karl Storz SE \& Co. KG, Tuttlingen, Germany) at a resolution of $1920 \times 1080$, and the \textit{Image1S Connect TC200} (Karl Storz SE \& Co. KG, Tuttlingen, Germany) at a resolution of $3840 \times 2160$. The \textit{AIDA} (Karl Storz SE \& Co. KG, Tuttlingen, Germany) recording system was used to record the surgeries at $25 fps$. Similar to the simulator data, relevant frames were extracted from the captured videos and then manually labelled using the \textit{labelme} software. The prepared dataset contains a total of $2,376$ frames ($1,188$ stereo-pairs), with approx. $23,900$ labelled suture landmarks. The intra-operative dataset was organised into $4$ folds, with each surgery comprising one fold and leave-one-out cross validation being performed. A separate test set is prepared in the intra-operative domain comprising of data from $5$ unseen surgeries. The dataset is made up of $100$ images ($50$ stereo-pairs) from each surgery, adding up to a total of $500$ images, with approx. $5,500$ landmark labels.

To provide further insights into the suggested dataset, we show some statistics over it with respect to light intensity distribution, number of sutures per images and difficulties of annotations. The difficulty levels of labeling the suture landmarks of an image were marked by the labellers, based on the spatial orientation, visibility of suture, and nature of the scene. The distribution of the difficulty across the datasets in both the domains, is presented in Figure \ref{fig:data_analysis} (a). 
Secondly, the histogram distribution for the number of sutures across the images of the training dataset is provided in Figure \ref{fig:data_analysis} (b). Furthermore, the distributions of the average light intensity, captured in the \textit{Value}-channel of the \textit{HSV} colour-space across each fold of the simulator and intra-operative datasets, are shown in Figure \ref{fig:data_analysis} (c). It can be seen that strong reflections from various light sources (head lamp, endoscopic light) appear as peaks in most histograms and that the histograms differ for each fold and for each domain. Finally, a \textit{Sankey} relational diagram is presented in Figure \ref{fig:data_analysis} (d), that represents the proportion of the training dataset in the intra-operative domain in relation to the aforementioned attributes of the intra-operative dataset.

\subsection{Training}
\label{sub:training}

\subsubsection{Landmark detection}
\label{subsub:landmark_detection}
A landmark detection network, with an encoder-decoder architecture based on the UNet \cite{Ronneberger_Unet2015} is used to detect suture landmarks from endoscopic images. The network is made up of four downsampling and four upsampling blocks. The upsampling blocks perform bilinear interpolation. Dropout is used after a convolutional block, with values changing step-wise from $p \in [0.3, 0.5]$, with $0.3$ at the outermost layers and $0.5$ at the bottleneck layer. Each convolutional block comprises a convolutional layer, followed by a \textit{ReLU} activation function and a \textit{Batchnorm} layer. There is a $1 \times 1$ convolutional at the end of the network, followed by a \textit{sigmoid} activation layer.
A spatial \textit{Gaussia}n filter layer of kernel size $(3\times3)$ and $\sigma=1$, which is differentiable is applied to the output of the \textit{Sigmoid} layer, followed by a spatial convolutional differentiable $2D$ \textit{Soft-Argmax} layer, with a kernel size of $(3\times3)$. The outputs from the \textit{Sigmoid} layer and the \textit{Soft-Argmax} layer are both learned to optimise the similarity function $(MSE + 1 - DICE)$ with the ground-truth \textit{Gaussian} heatmap. The \textit{Gaussian} and \textit{Soft-Argmax} layers are implemented using the \textit{Kornia} \cite{eriba2019kornia} library.

The input images have a resolution of $512 \times 288$, maintaining the aspect ratio of the original images. For each labeled suture point in an image, a $2D$ \textit{Gaussian} is computed centered around the landmarks, as described in \ref{sub:landmark_detection_methods}, with $\sigma=2$. The labeled point mask, created at a resolution $512 \times 288$ is provided along with the image, to obtain predictions of the same resolution. A color augmentation is performed on the images before training with a probability of $0.5$. The color augmentation is composed of shifting the brightness values in range $\pm 0.2$, the contrast in range $[0.3, 1.5]$, saturation in range $[0.5, 2]$, and hue in range $\pm 0.1$. Additional augmentation is performed on both the images and masks, each with a probability of $0.5$, namely: rotation in range $\pm 60^\circ$, translation in range $\pm 10\%$, shear in range $\pm 0.1$, random horizontal and vertical flip. The \textit{Albumentations} \cite{DBLP:journals/information/BuslaevIKPDK20} library is used to perform the data augmentations. The network is trained using a standard \textit{Adam} optimizer, with a batch size of $32$ and a learning rate of $0.001$. The learning rate is reduced by a factor of $0.1$, if it plateaus for a period of $10$ epochs. The \textit{PyTorch} library \cite{paszke2017automatic} was used for training the models. The training was performed on either of \textit{Nvidia Quadro P6000}, \textit{TITAN RTX}, or \textit{TITAN V}. 

\subsubsection{Image translation networks}
\label{subsub:image_translation_networks}
The unpaired image-to-image translation network is based on the CycleGAN from Zhu et al. \footnote{https://github.com/aitorzip/PyTorch-CycleGAN} \cite{zhu_unpaired_2020}. The two generators used, $G_{sim2or}$ and $G_{or2sim}$ are ResNet \cite{he_deep_2016} based architectures. Each generator is composed of the following sequence: A convolutional block, $2$ downsampling blocks, $6$ residual blocks with $32$ filters, two upsampling blocks, and an output convolutional layer.
Each convolutional block comprises of a convolutional layer, followed by an \textit{instance normalization} layer, and a \textit{ReLU} activation layer. The convolutional layers have filter size $3$, except in the input and output layers that have a filter size of $7$. The downsampling and upsampling blocks have strided convolutions, and the final convolutional layer has a $tanh$ activation. The two discriminators $D_{or}$ and $D_{sim}$ each comprise of a sequence of five convolutional layers with a filter size of four, and a \textit{Leaky ReLU} activation function. The layers two, three and five additionally contain an \textit{instance normalization} layer. 

For the CycleGAN baseline model, the images are input at a resolution of $512 \times 288$. The images are standardized in the range $\mu=0.5$, and $\sigma=0.5$ before performing data augmentation. The following data augmentation is performed on the input image, each with a probability of $0.5$: Rotation in range $\pm 60^\circ$, horizontal flip, and vertical flip. $3$ instances of the standard \textit{Adam} optimizer are used for training the model: One for the generators and one each for the $2$ discriminators $D_{or}$ and $D_{sim}$. The network is trained for $60$ epochs, with a learning rate of $0.0002$ and a batch size of $8$. The corresponding folds of the datasets from each domain are used to train the model. The \textit{DetCycleGAN} is trained with the same hyper-parameters as the baseline. 
The weights of the pre-trained landmark detection networks from each domain are loaded and kept fixed during training. The training of the CycleGAN models were performed on either of \textit{Nvidia Quadro P6000} or the \textit{TITAN RTX}. 

\begin{figure}[t]
    \centering
    \includegraphics[width=0.5\textwidth]{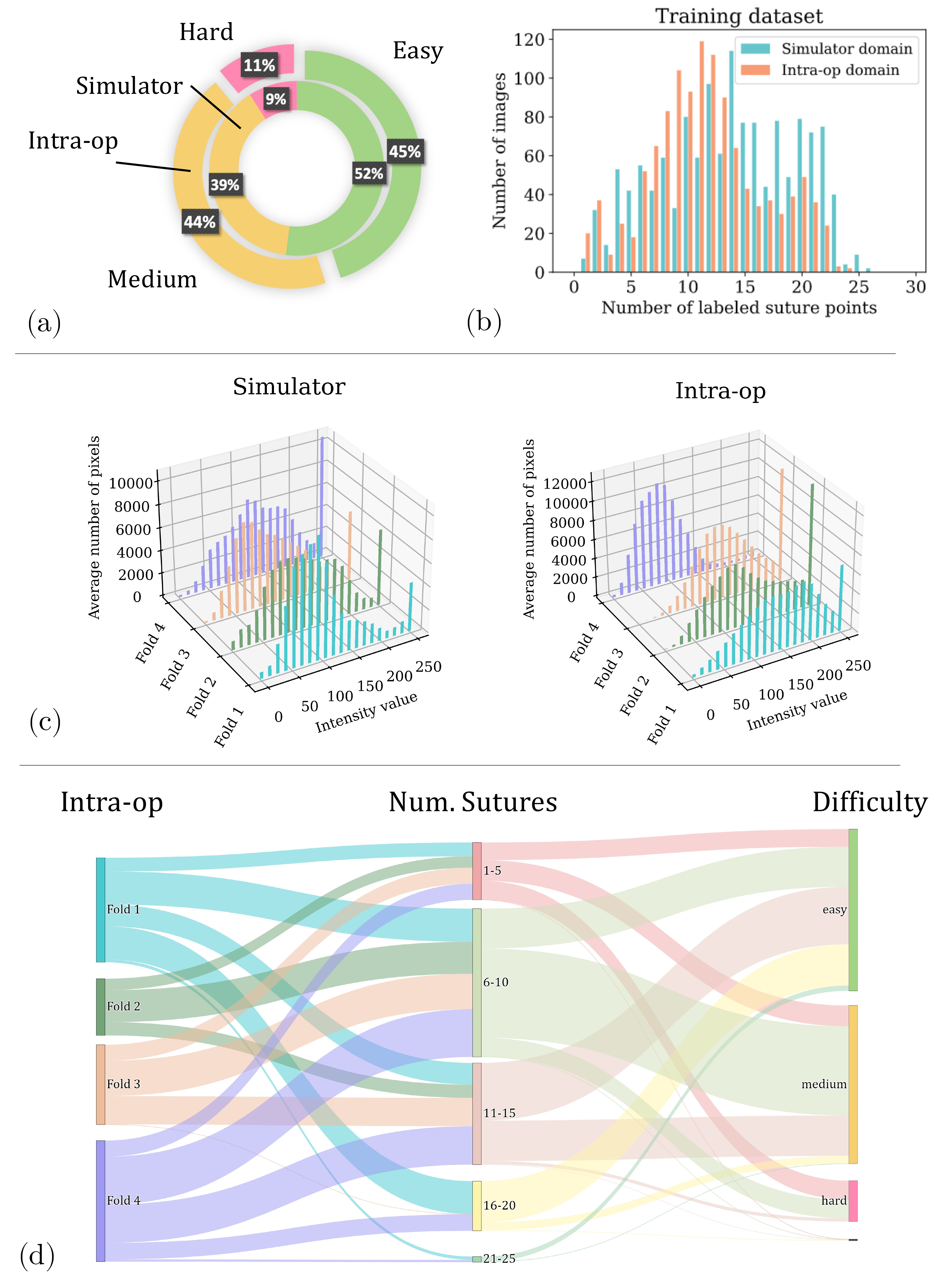}
    \caption{(a) breakup of the difficulty of labeling the sutures. (b) distribution of the number of sutures for the training dataset in both the domains. (c) distribution of average light intensity in both domains. (d) a Sankey relational diagram for the training data in the intra-op domain. The flow between the nodes represents the proportion of the dataset that corresponds to the two respective categories.}
    \label{fig:data_analysis}
\end{figure}

\subsection{Evaluation}
\label{sub:evaluation}

In order to evaluate the landmark detection network, the predicted masks are compared with the ground truth masks by computing the distance between the centers of mass of the points. If the computed distance is less than a threshold, the point is considered to be detected. If multiple points are detected for a ground-truth point, then the point with the closest distance to the ground-truth point is assigned to the predicted point. In our case, a threshold of $6$ pixels \cite{DBLP:conf/bildmed/SternSRKKSWE21} is employed. Based on this criteria, the true positives (TP), false positives (FP) and the false negatives (FN) are computed. The precision or the \textit{Positive Predicted Value (PPV)} is computed as ${TP}/{(TP+FP)}$, and the sensitivity or the \textit{True Positive Rate (TPR)} is computed as ${TP}/{(TP+FN)}$. 
Finally, the $F_1$ score is reported by taking the harmonic mean of $PPV$ and $TPR$, as $F_1 = (PPV \times TPR) / (PPV + TPR)$.

For the image translation network, multiple metrics such as the \textit{Fr\'echet Inception Distance (FID)} \cite{Heusel2017GANsTB} and the \textit{Single Image Fr\'echet Inception Distance (SIFID)} \cite{shaham_singan_2019} have been proposed to evaluate the quality of image-to-image translation. However, these metrics measure the similarity between two distributions and thereby indicate the similarity in the style between both the domains. Such a metric may not be well suited to the task of rating the consistency in transfer of fine-grained information such as the suture points. 
In our case, we compare the performance of the suture detection network trained in the intra-operative domain, $Det_{or}$, on the fake \textit{OR} data $x'_{or}$ generated by the CycleGAN baseline, and the \textit{DetCycleGAN} variants. 

Additionally, to show that the whole sutures and not just the entry and exit points are consistently translated, we perform a tailored analysis of the fake images. The fake intra-operative images generated from the following models are chosen for comparison in this study: The baseline CycleGAN model, the Variant $1$ of the \textit{DetCycleGAN} model, with $\alpha_1 = 1, \alpha_2 = 1$, and the Variant $2$ of the \textit{DetCycleGAN} model, with $\alpha_1 = 1, \alpha_2 = 0$. The suture segments in the fake generated intra-operative domain were labeled by an annotator who is blinded to the ground truth image from the simulator domain. 
A total of $96$ images, comprising of $8$ images from each fold, coming from $3$ different experiments, were presented in a random order to an annotator. 
Similarity metrics of $MSE$ and the $DICE$ score are then computed between the suture masks of the simulator and the translated image.

Finally, we show that the generated images $x'_{or}$, in the \textit{OR} domain can be used to improve the performance of the landmark detection network. The landmark detection network $Det_{or}$ is trained from scratch on the fused dataset comprising of real $x_{or}$ and fake $x'_{or}$ intra-operative images. The trained model is evaluated on the additional test set prepared in the intra-operative domain. The complete evaluation strategy is depicted in Figure \ref{fig:process}.

\subsection{Ablations and Variants}
\label{sub:ablations_and_variants}
We present experiments with the CycleGAN without the integration of the landmark detection networks $Det_{dom}$, $dom \in \{sim, or\}$, which comprises the baseline. Additionally, two variants of the \textit{DetCycleGAN} (cf. Equation \ref{eq:final_objective}) are presented. 
In Variant $1$, both $\mathcal{L}_{Det'}$ and $\mathcal{L}_{Det''}$ from Equation \ref{eq:final_objective} are applied. Here, we present experiments with $3$ different loss weighting strategies, with the following combinations of $(\alpha_1, \alpha_2)$: $(1, 1), (0.5, 1), \text{~and~} (1, 0.5)$. In Variant $2$, only the $\mathcal{L}_{Det'}$ is applied, i.e. $\alpha_1 = 1 \text{~and~} \alpha_2=0$ in Equation \ref{eq:final_objective}. Moreover, another experiment was performed by additionally including a \textit{cross-domain consistency loss} in the form of \textit{Mean Squared Error (MSE)} between two pairs of masks. First, those predicted on $x'_{or}$ and $x''_{sim}$, and second, $x'_{sim}$ and $x''_{or}$, similar to that proposed by Chen \textit{et al.} \cite{chen_crdoco_2020}.

For means of comparison to state-of-the-art, we conducted a comparison between the CycleGAN architecture and the Fast \textit{Contrastive Unpaired Translation} (Fast CUT) method \cite{park2020cut} for unpaired image-to-image translation. Additionally, we perform an experiment where we adapt the concept of the semantic loss from a domain adaptation model CyCADA \cite{hoffman_cycada_2017}, where we use the landmark detection networks in the simulator and intra-operative domains as noisy labellers for their respective corresponding domains, and force the predicted outputs of a domain to be consistent before and after translation. This can be expressed as

\begin{equation}
    \begin{aligned}
        \mathcal{L}_{sem} &= \mathcal{L}_{Det}(Det_{sim}(x_{or}), Det_{sim}(x'_{sim})) \\ 
        &+  \mathcal{L}_{Det}(Det_{or}(x_{sim}), Det_{or}(x'_{or}))
    \end{aligned}.
\label{eq:semantic_loss}
\end{equation}

\section{Results}
\label{sec:results}

\subsection{Detection Networks: Domain Comparison} 
\label{sub:detection_network_results}
Firstly, the landmark detection network $Det_{or}$ is trained and evaluated on the respective folds of the intra-operative domain and the results of cross-validation are presented in Table \ref{tab:results1}. Similarly, the other $Det_{sim}$ network is cross-validated on the simulator domain. 
The performance of the landmark detection model is better in the simulator domain than in the intra-operative domain (mean \textit{PPV} $+14.41$,  mean \textit{TPR} $+27.27$, mean $F_1$ $+0.2450$).
The intra-operative dataset contains high variability in illumination, white balance, pose, view angle, blood, and endoscopic artefacts such as occlusion (cf. Figure \ref{fig:data_analysis} (d)). This makes the dataset in the intra-operative domain challenging compared to the more homogeneous structure and scene composition of the simulator domain. 

  \begin{table}[htbp]
  \centering
  \resizebox{\linewidth}{!}{%
    \begin{tabular}{llccccc} \hline 
    \multicolumn{7}{c}{Cross-validation of landmark detection on $Sim$ and $OR$ real data} \\ \hline
      Model & Metric & $f1$ &  $f2$ & $f3$ & $f4$ & $\mu \pm \sigma$ \\ \hline
      $Det_{or}(x_{or})$ & PPV & $75.13$  & $60.89$  & $60.28$  & $77.79$  & $68.52\pm8.00$ \\ 
                         & TPR & $43.34$  & $34.51$  & $53.84$  & $30.94$  & $40.66\pm8.85$ \\ 
                         & $F_1$ & $0.5497$ & $0.4405$ & $0.5688$ & $0.4427$ & $0.5004\pm0.06$ \\ \hline
      
      $Det_{sim}(x_{sim})$ & PPV & $83.26$  & $83.87$  & $88.36$  & $76.21$  & $\mathbf{82.93\pm4.35}$ \\ 
                           & TPR & $75.09$  & $69.49$  & $62.56$  & $64.59$  & $\mathbf{67.93\pm4.84}$ \\ 
                           & $F_1$ & $0.7896$ & $0.7601$ & $0.7325$ & $0.6992$ & $\mathbf{0.7454\pm0.03}$ \\ \hline
      \end{tabular}%
}
  \caption{Cross-validation of landmark detection performed across two domains. The networks $Det_{or}$ and $Det_{sim}$ are trained and evaluated on folds $f1-f4$ from the intra-operative domain $x_{or}$, and the simulator domain $x_{sim}$, respectively.}
  \label{tab:results1}
  \end{table}

\subsection{Image-to-Image translation: Baseline CycleGAN vs. CycleGAN with integrated detection losses}
\begin{figure}[t]
    \centering
    \includegraphics[width=0.5\textwidth]{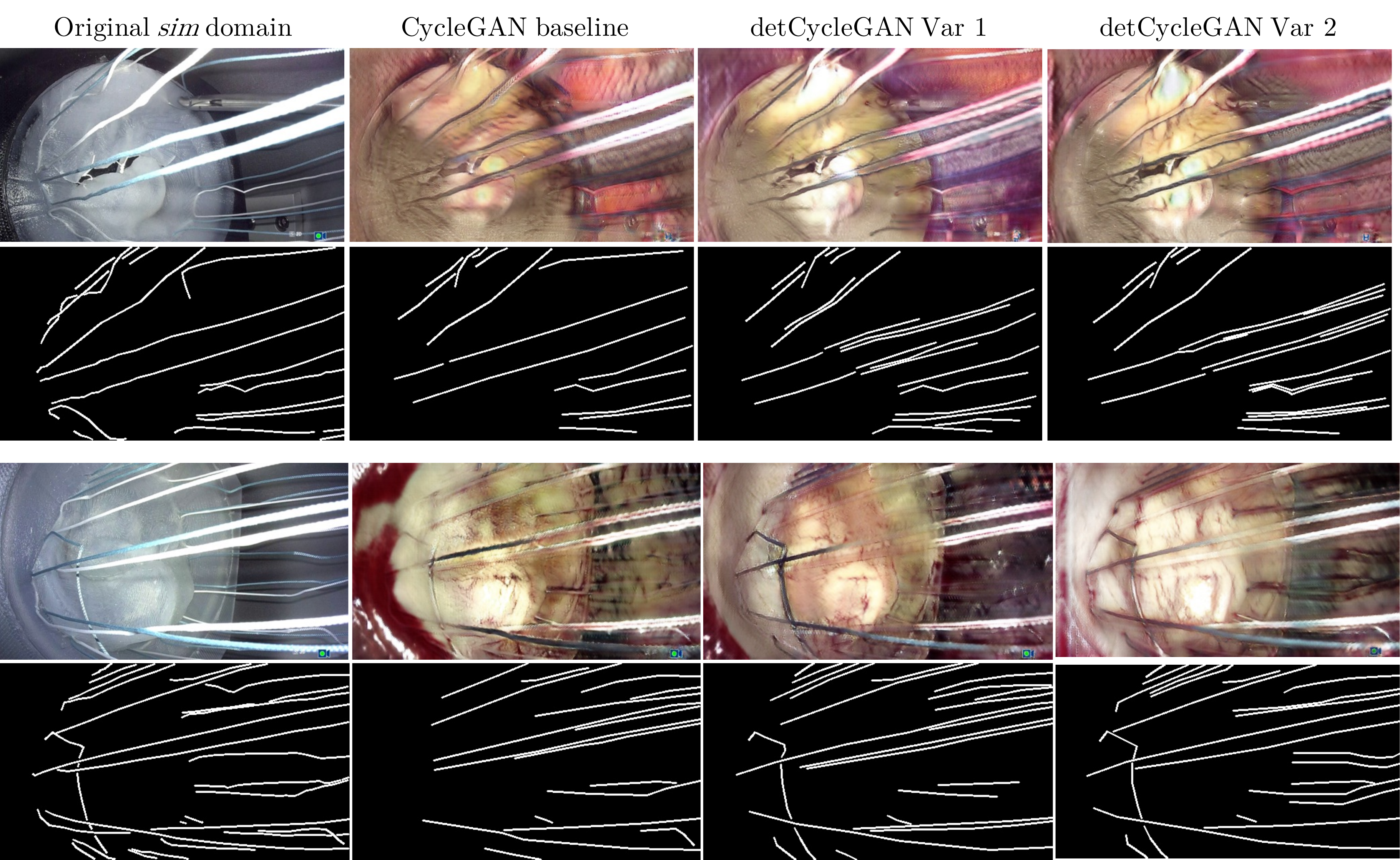}
    \caption{Samples of images generated from various models, in the fake intra-operative domain $x'_{or}$, generated by $G_{sim2or}$; with the respective suture annotations.}
    \label{fig:suture_labels}
\end{figure}

\begin{figure}[t]
    \centering
    \includegraphics[width=0.5\textwidth]{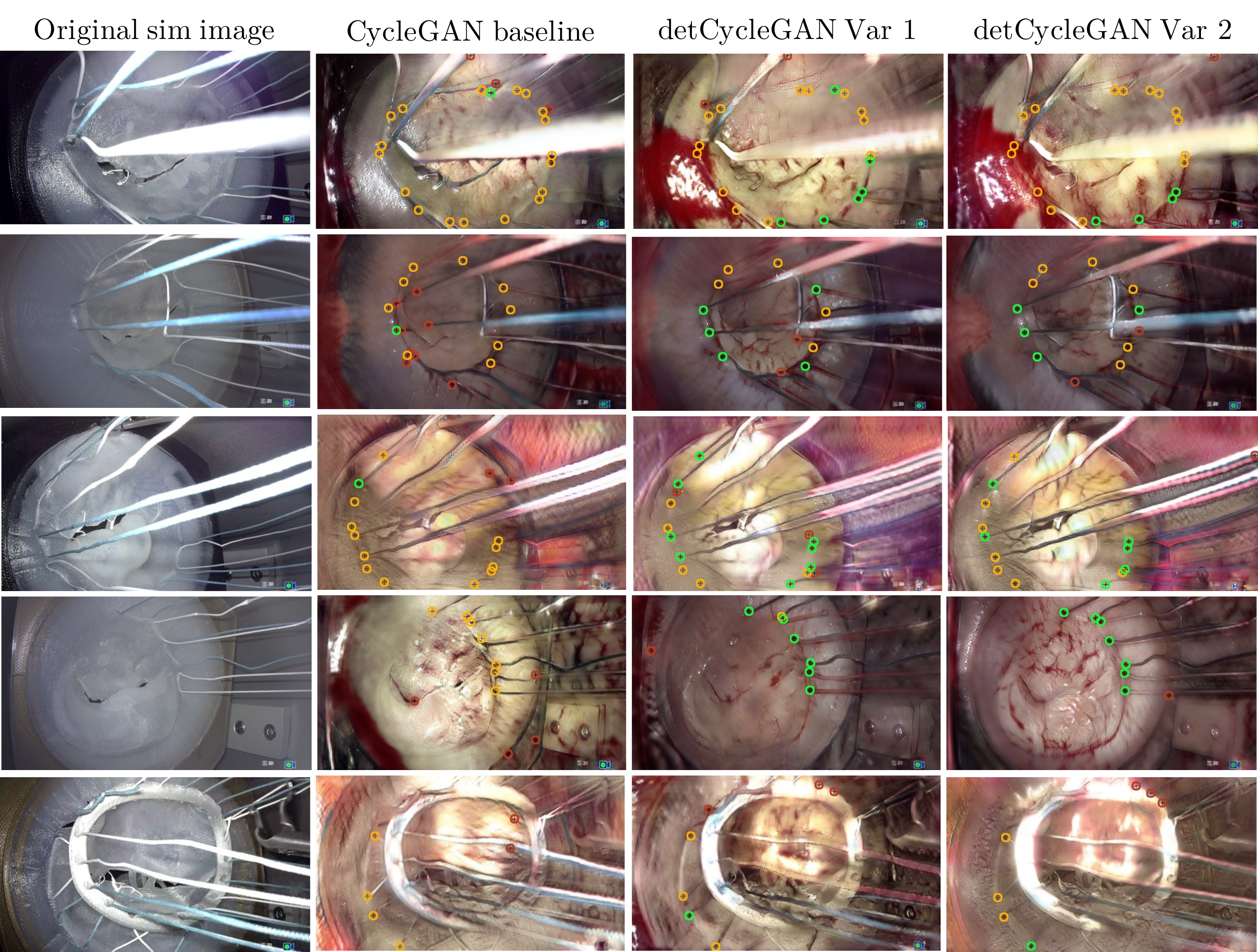}
    \caption{Samples of generated images from various folds, in the fake intra-operative domain $x'_{or}$, generated by $G_{sim2or}$ with annotations of the detected landmarks. The colored circles indicate true positives (green), false positives (red) and false negatives (orange).}
    \label{fig:annotated}
\end{figure}

Secondly, the landmark detection model $Det_{or}$ that is trained on folds of the intra-operative data $x_{or}$ is predicted on the data generated by the CycleGAN baseline and the \textit{DetCycleGAN} variants. The results are presented in Table \ref{tab:results2}. The landmark detection model $Det_{or}$ performs better on the data generated by the \textit{DetCycleGAN} and its variants, as compared to the baseline, indicating that the synthesis of sutures is better preserved. Specifically, the Variant $1$ with $\alpha_1=1, \alpha_2=1$ is best performing model. Indeed, we see that the entire suture, not only the parts closer to the valve, is captured in the target domain. This finding is explicitly backed up by the results presented in Table \ref{tab:results4} and in Fig. \ref{fig:suture_labels}. The Variants $1$ and $2$ of the proposed \textit{DetCycleGAN} are able to translate the sutures in a more consistent manner, in comparison to the baseline.
Furthermore, sutures at wrong positions, and therefore false positives, are reduced. Example comparisons of generated images and landmark prediction are shown in Figure \ref{fig:suture_labels} and Figure \ref{fig:annotated}. Surprisingly, the \textit{DetCycleGAN} variant with the \textit{cross-domain consistency loss}, and the adapted \textit{Semantic loss} generated fake intra-operative samples that are visually worse when compared to the baseline CycleGAN model. Therefore, we did not include it for quantitative comparison.

   \begin{table}[htbp]
  \centering
  \resizebox{\linewidth}{!}{%
    \begin{tabular}{llccccccc} \hline 
    \multicolumn{9}{c}{Cross-validation of landmark detection performance on fake $OR$ data from variants} \\ \hline
      Metric & Experiment & $\alpha_1$ & $\alpha_2$ & $f1$ &  $f2$ & $f3$ & $f4$ & $\mu \pm \sigma$ \\ \hline
      PPV & $\mathcal{L}_{CycleGAN}$ (Baseline) & & & $14.14$ & $14.51$ & $18.20$ & $8.26$ & $13.78\pm3.56$ \\ 
      & $\mathcal{L}_{DetCycleGAN}$ (Var $1$)  & $1$ & $1$ & $76.02$ & $70.99$ & $79.95$ & $73.42$ & $\mathbf{75.10\pm3.32}$ \\
      & $\mathcal{L}_{DetCycleGAN}$ (Var $1$)  & $1$ & $0.5$ & $76.70$ & $63.23$ & $84.70$ & $70.30$ & $73.73\pm7.92$ \\
      & $\mathcal{L}_{DetCycleGAN}$ (Var $1$)  & $0.5$ & $1$ & $69.39$ & $61.95$ & $83.39$ & $64.64$ & $69.84\pm8.26$ \\
      & $\mathcal{L}_{DetCycleGAN}$ (Var $2$)  & $1$ & $0$ & $77.39$ & $63.29$ & $77.47$ & $70.69$ & $72.21\pm5.84$ \\
      & \textit{Fast CUT} \cite{park2020cut} & & & $9.20$ & $15.30$ & $36.60$ & $5.94$ & $16.76\pm11.94$ \\\hline
      
      TPR & $\mathcal{L}_{CycleGAN}$ (Baseline) & & & $4.59$ & $4.39$ & $8.27$ & $1.65$ & $4.73\pm2.35$ \\ 
      & $\mathcal{L}_{DetCycleGAN}$ (Var $1$)  & $1$ & $1$ & $42.71$ & $41.97$ & $46.10$ & $39.77$ & $\mathbf{42.64\pm2.28}$ \\
      & $\mathcal{L}_{DetCycleGAN}$ (Var $1$)  & $1$ & $0.5$ & $38.79$ & $39.24$ & $54.17$ & $34.89$ & $41.77\pm7.35$ \\
      & $\mathcal{L}_{DetCycleGAN}$ (Var $1$)  & $0.5$ & $1$ & $30.87$ & $37.47$ & $49.67$ & $31.19$ & $37.30\pm7.61$ \\
      & $\mathcal{L}_{DetCycleGAN}$ (Var $2$)  & $1$ & $0$ & $39.02$ & $33.80$ & $51.73$ & $44.10$ & $42.16\pm6.62$ \\
      & \textit{Fast CUT} \cite{park2020cut} & & & $2.01$ & $2.28$ & $2.34$ & $0.56$ & $1.80\pm0.73$ \\\hline
      
      $F_1$ & $\mathcal{L}_{CycleGAN}$ (Baseline) & & & $0.0693$ & $0.0674$ & $0.1138$ & $0.0275$ & $0.0695\pm0.03$ \\ 
      & $\mathcal{L}_{DetCycleGAN}$ (Var $1$)  & $1$ & $1$ & $0.5469$ & $0.5275$ & $0.5848$ & $0.5160$ & $\mathbf{0.5438\pm0.03}$ \\
      & $\mathcal{L}_{DetCycleGAN}$ (Var $1$)  & $1$ & $0.5$ & $0.5152$ & $0.4842$ & $0.6608$ & $0.4664$ & $0.5317\pm0.08$ \\
      & $\mathcal{L}_{DetCycleGAN}$ (Var $1$)  & $0.5$ & $1$ & $0.4273$ & $0.4670$ & $0.6226$ & $0.4208$ & $0.4844\pm0.08$ \\
      & $\mathcal{L}_{DetCycleGAN}$ (Var $2$)  & $1$ & $0$ & $0.5188$ & $0.4407$ & $0.6203$ & $0.5432$ & $0.5308\pm0.06$ \\
      & \textit{Fast CUT \cite{park2020cut}} & & & $0.0330$ & $0.0397$ & $0.0440$ & $0.0102$ & $0.0317\pm0.01$ \\\hline
      \end{tabular}%
}
  \caption{Performance of landmark detection on the images $x'_{or}$ generated by the extended unpaired image-to-image translation framework and ablated variants. The fixed network $Det_{or}$ was previously trained on $x_{or}$ images (cf. Table \ref{tab:results1}), and is evaluated here on the evaluation splits of fake images.}
  \label{tab:results2}
  \end{table}

 \subsection{Detection Networks: Fusion of intra-operative training datasets with generated samples}
 \label{sub:fusion}
 Besides generating suture-consistent image-to-image translation, we assess if the generated images can in turn be used to further improve performance on the landmark detection task. The landmark detection network $Det_{or}$ is trained from scratch, together on the real intra-operative images $x_{or}$ and the fake generated images of the target domain $x'_{or}$. This model is now evaluated on an additional unseen intra-operative test set (not used in the experiments before).
 From Table \ref{tab:results3}, it can be observed that when retraining the landmark detection network with the fused dataset from the CycleGAN baseline, the performance decreases compared to just using the real intra-operative data (mean $F_{1}$ $-0.0104$). When retraining with the fused dataset from the proposed \textit{DetCycleGAN} variants, the performance on the external test set is better when compared to using just the real intra-operative data (mean $F_{1}$ $+0.0518$ for \textit{DetCycleGAN} Variant $1$, and mean $F_{1}$ $+0.0581$ for \textit{DetCycleGAN} Variant $2$; cf. Table \ref{tab:results3}).

\begin{table}[htbp]
  \centering
  \resizebox{\linewidth}{!}{
    \begin{tabular}{llccccc} \hline 
    \multicolumn{7}{c}{Comparison of similarity between labeled sutures in $x'_{or}$ and $x_{sim}$} \\ \hline
      Metric & Experiment & $f1$ &  $f2$ & $f3$ & $f4$ & $\mu$ $\pm$ $\sigma$ \\ \hline
      MSE & $\mathcal{L}_{CycleGAN}$ (Baseline) & $0.0638$ & $0.0547$ & $0.0608$ & $0.0421$ & $0.0554\pm0.0083$ \\ 
      & $\mathcal{L}_{DetCycleGAN}$ (Var $1$)   & $0.0554$ & $0.0461$ & $0.0602$ & $0.0421$ & $ \mathbf{0.0510\pm0.0072}$ \\
      & $\mathcal{L}_{DetCycleGAN}$ (Var $2$)   & $0.0683$ & $0.0467$ & $0.0462$ & $0.0484$ & $0.0524\pm0.0092$ \\\hline
      
      Dice & $\mathcal{L}_{CycleGAN}$ (Baseline) & $0.4233$ & $0.4474$ & $0.4518$ & $0.5108$ & $0.4583\pm0.0322$ \\ 
      & $\mathcal{L}_{DetCycleGAN}$ (Var $1$)   & $0.4905$ & $0.5368$ & $0.4794$ & $0.5293$ & $ \mathbf{0.5090\pm0.0245}$ \\
      & $\mathcal{L}_{DetCycleGAN}$ (Var $2$)   & $0.4255$ & $0.5320$ & $0.5839$ & $0.4744$ & $0.5040\pm0.0596$ \\\hline
      \end{tabular}%
    }
  \caption{Similarity of sutures labeled from the predicted output $x'_{or}$ of the model variants with the original corresponding image $x_{sim}$ from the simulator domain.}
  \label{tab:results4}
  \end{table}
  
    \begin{table}[htbp]
  \centering
  \resizebox{\linewidth}{!}{
    \begin{tabular}{llccccc} \hline 
    \multicolumn{7}{c}{Results of $Det_{or}$ on additional separated $OR$ test set} \\ \hline
      Metric & Experiment & $f1$ &  $f2$ & $f3$ & $f4$ & $\mu$ $\pm$ $\sigma$ \\ \hline
      PPV & only real $x_{or}$                                            & $67.83$ & $78.65$ & $78.09$ & $81.79$ & $76.59\pm5.25$  \\ 
      & $x_{or} + x'_{or}$ from $\mathcal{L}_{CycleGAN}$ (Baseline)       & $70.62$ & $74.59$ & $67.36$ & $69.91$ & $70.62\pm2.59$ \\
      & $x_{or} + x'_{or}$ from $\mathcal{L}_{DetCycleGAN}$ (Var $1$)    & $83.16$ & $77.33$ & $76.82$ & $75.62$ & $\mathbf{78.23\pm2.91}$  \\
      & $x_{or} + x'_{or}$ from $\mathcal{L}_{DetCycleGAN}$ (Var $2$) & $68.46$ & $70.76$ & $72.88$ & $70.96$ & $70.77\pm1.57$  \\ \hline
      
      TPR & only real $x_{or}$                                            & $34.12$ & $30.64$ & $29.37$ & $25.38$ & $29.88\pm3.13$ \\
      & $x_{or} + x'_{or}$ from  $\mathcal{L}_{CycleGAN}$ (Baseline)      & $21.07$ & $37.33$ & $31.47$ & $29.63$ & $29.88\pm5.82$ \\
      & $x_{or} + x'_{or}$ from $\mathcal{L}_{DetCycleGAN}$ (Var $1$)    & $29.83$ & $33.71$ & $37.87$ & $37.37$ & $34.70\pm3.24$ \\
      & $x_{or} + x'_{or}$ from $\mathcal{L}_{DetCycleGAN}$ (Var $2$) & $37.27$ & $38.80$ & $35.52$ & $36.27$ & $\mathbf{36.97\pm1.23}$ \\ \hline
      
      $F_1$ & only real $x_{or}$                                         & $0.4540$ & $0.4410$ & $0.4268$ & $0.3874$ & $0.4273\pm0.02$ \\
      & $x_{or} + x'_{or}$ from  $\mathcal{L}_{CycleGAN}$ (Baseline)     & $0.3246$ & $0.4976$ & $0.4290$ & $0.4162$ & $0.4169\pm0.06$ \\
      & $x_{or} + x'_{or}$ from $\mathcal{L}_{DetCycleGAN}$ (Var $1$)   & $0.4391$ & $0.4696$ & $0.5073$ & $0.5002$ & $0.4791\pm0.03$ \\
      & $x_{or} + x'_{or}$ from $\mathcal{L}_{DetCycleGAN}$ (Var $2$) & $0.4827$ & $0.5012$ & $0.4776$ & $0.4800$ & $\mathbf{0.4854\pm0.01}$ \\ \hline
      \end{tabular}
}
  \caption{The trained generator networks from the folds $f1-f5$ in Table \ref{tab:results2} are applied to the simulation domain to generate $x'_{or}$. A $Det_{or}$ network is then trained from scratch, on fused dataset comprising real and fake \textit{OR} data. This network is predicted on the unseen intra-operative test set consisting $500$ frames from $5$ surgeries.}
  \label{tab:results3}
  \end{table}

\section{Discussion}
\label{sec:discussion}
For complex surgeries like mitral valve repair that have a steep learning curve, generative models play a crucial role in improving the realism of surgical simulators and thereby the quality of surgical training. In this context, landmark detection tasks such as suture detection involve identifying an unknown number of sutures in each frame, which potentially lack a clear structural relation. Besides, due to the nature of data scarcity and data privacy in the surgical domain, it is particularly challenging to acquire intra-operative data in comparison to other imaging modalities. Furthermore, surgical data is highly heterogeneous, due to varying lighting conditions, acquisition angles, and the number and type of objects in the scene \cite{sharan_domain_2020}.  It can indeed be observed that the detector network $Det_{sim}$ is able to better learn the distribution of the simulator domain than the $Det_{or}$ network learns the distribution of the intra-operative domain (mean \textit{PPV} $+14.41$,  mean \textit{TPR} $+27.27$, mean $F_1$ $+0.2450$; cf. Table \ref{tab:results1}). Moreover, the \textit{PPV} and \textit{TPR} values show that the predictions comprise of a larger number of false negatives, as compared to false positives.

\begin{figure}[t]
    \centering
    \includegraphics[width=0.5\textwidth]{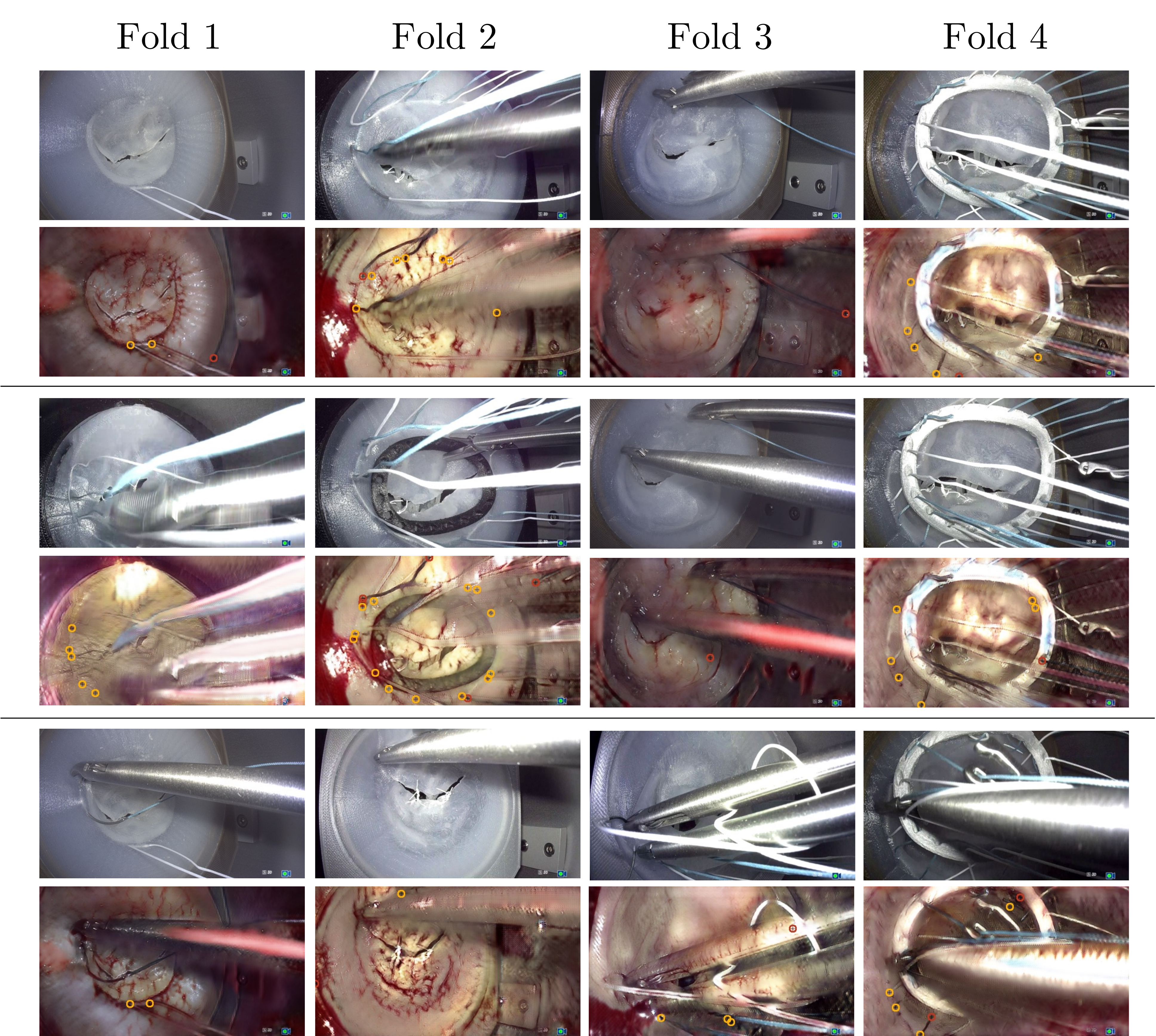}
    \caption{Samples of the failure cases of the \textit{DetCycleGAN} Variant $1$, along with the corresponding images in the Simulator domain: $3$ random samples from the predictions were no \textit{True Positives} are shown.}
    \label{fig:failure_cases}
\end{figure}

\begin{figure}[t]
    \centering
    \includegraphics[width=0.5\textwidth]{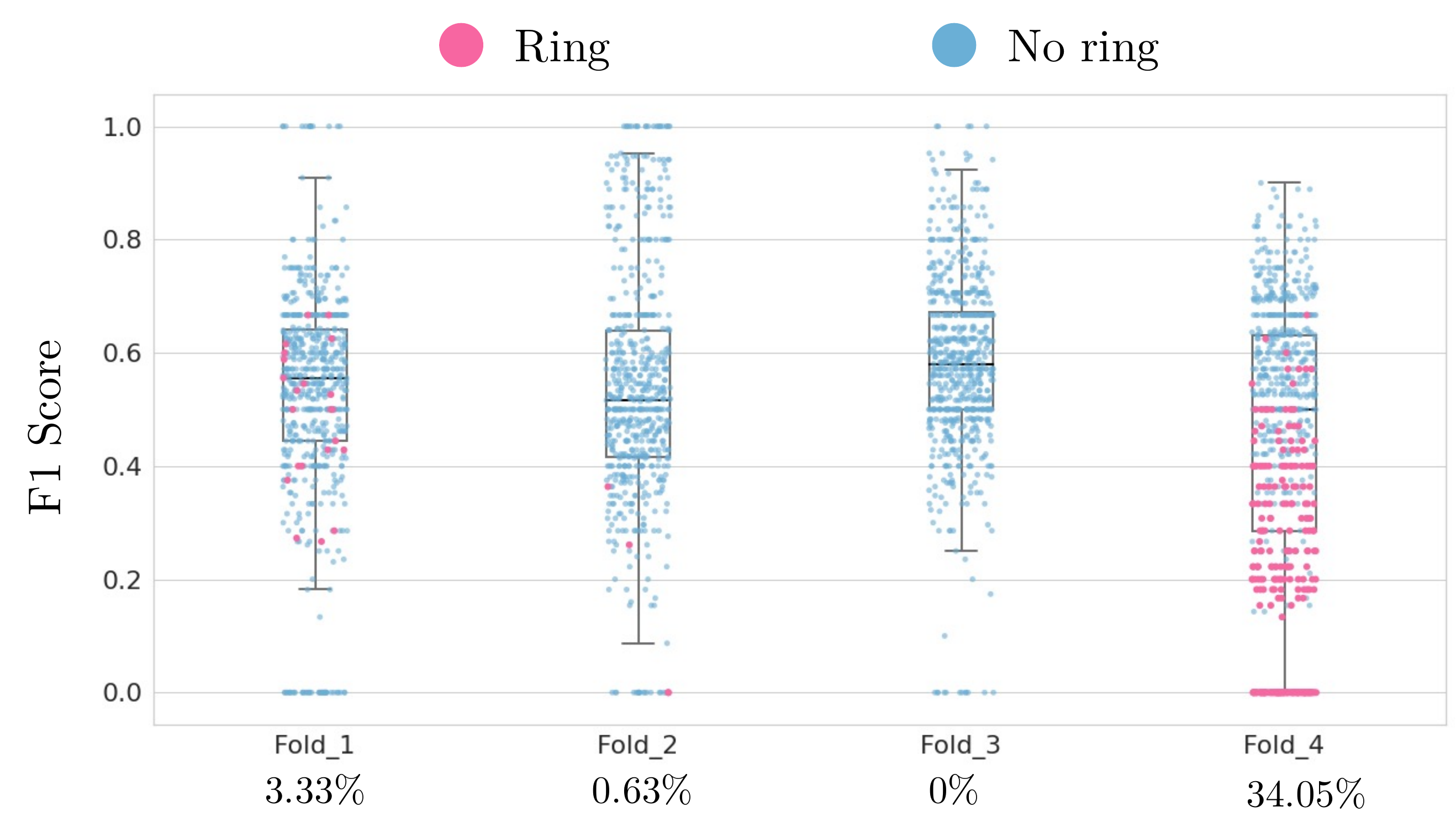}
    \caption{Distribution of the $F_1$ score across predicted datapoints of the \textit{DetCycleGAN} Variant $1$, for each fold of the Simulator domain transformed to the fake intra-operative domain. The percentage of images that contain an annuloplasty ring are shown below the fold numbers.}
    \label{fig:stripplot}
\end{figure}

Image-to-image translation networks struggle to generate images that are physically consistent across the source and target domains. 
To tackle this problem, multiple approaches from the literature suggest a combination of task networks and CycleGAN \cite{lin_multimodal_2020}, \cite{piao_semi-supervised_2019}. In endoscopy, Luengo \textit{et al.} \cite{Luengo2018} proposed a style transfer approach named \textit{SurReal} to enhance the realism of simulated data. 
In contrast to task-combined image translation networks that are typically used for unsupervised domain adaptation, our objective is not to learn the target labels from a labeled source dataset, but to translate images in a physically plausible manner and further use the translated images to improve the performance of landmark detection.
Our approach uses pre-trained task networks, which back-propagate the errors into the respective generator networks, such that their gradients are adjusted. This enforces suture-preserving transformations from the source to target domain. However, the reinforcement added by integrating the task network is dependent on the landmark detection accuracy of the task network. 
We demonstrate that the performance of the \textit{DetCycleGAN} variants outperform the baseline image translation model. In this work, we proposed  \textit{DetCycleGAN} Variant $1$, which uses a \textit{detection consistency loss} in both the fake and recovered domains ($\mathcal{L}_{CycleGAN} + \mathcal{L}_{Det'} + \mathcal{L}_{Det''}$). An ablation of it is using this loss only in the fake domain (Variant $2$, $\mathcal{L}_{CycleGAN} + \mathcal{L}_{Det'}$). Additionally, we present $3$ loss weighting strategies for the Variant 1 architecture.  Table \ref{tab:results2} reports prediction performance of the intra-operative landmark detection network on the translated fake intra-operative images, the Variant $1$ of the \textit{DetCycleGAN} is the top-performing model (PPV $75.10$, TPR $42.64$, $F_1$ $0.5438$). 
Moreover, random samples of the images from Variant $1$ of the \textit{DetCycleGAN} where no \textit{True Positives} are predicted, are shown in Figure \ref{fig:failure_cases}. It can be seen that the cases comprise of underexposed images, motion blurring and the images containing the annuloplasty ring. Furthermore, we plot the $F_1$ score of all the predicted data points in Figure \ref{fig:stripplot} and we highlight the specific attribute \textit{presence of ring prosthesis}. It can be seen that fold $4$, which is the worst performing fold with respect to the $F_1$ score in Table \ref{tab:results2}, comprises of more samples where an annuloplasty ring is included.

Similarly, the analysis of the labeled sutures demonstrates that the sutures in the fake intra-operative images produced by the Variant $1$ of the \textit{DetCycleGAN} is closest to the ground truth (mean $MSE~0.0510\pm0.0083$, mean $Dice~0.5090\pm0.0245$) in comparison with the images generated by the baseline and Variant $2$. However, note that this evaluation method of labeling and comparing the centerlines of the sutures is still prone to some label inaccuracies. The metrics of $MSE$ and $DICE$ therefore provide a rough approximation  of how consistent the sutures are in the translated images compared to the ground truth.
Finally, we showed that the performance of the landmark detection network improves when trained on a fused dataset comprising of real and fake images from the intra-operative domain. Both the variants of the proposed \textit{DetCycleGAN} model perform better than just using only the real intra-operative data (mean $F_1$ $+0.0518$ for \textit{DetCycleGAN} Variant $1$, and mean $F_1$ $+0.0581$ for \textit{DetCycleGAN} Variant $2$; cf. Table \ref{tab:results3}).
This is also an interesting aspect towards tackling the problem of data privacy and data sharing in the surgical domain. The generated fake intra-operative datasets can be more easily shared in a privacy-preserving manner \cite{pfeiffer2019generating}.

\section{Conclusion}
In conclusion, we have shown that an unpaired image-to-image approach can be directly coupled to a landmark detection task to improve synthesis of foreground objects like sutures. Having the goal in mind of displaying such images in real-time to surgeons when they train a minimally-invasive procedure on a simulator, all foreground objects must be clearly visible; meaning that the consistency between source and target domain is of paramount importance. 
We have made the simulator and part of the intra-operative data available to the community within the scope of the AdaptOR MICCAI Challenge 2021 \cite{Engelhardt_AdaptOR}.


\bibliography{references.bib} 
\bibliographystyle{IEEEtran}

\end{document}